\def\year{2018}\relax
\documentclass[letterpaper]{article} 
\usepackage[noblocks]{authblk}
\usepackage{aaai18}  
\usepackage{times}  
\usepackage{helvet}  
\usepackage{courier}  
\usepackage{url}  
\usepackage{graphicx}  
\usepackage[T1]{fontenc} 

\usepackage{booktabs} 
\usepackage{subcaption} 
\usepackage{verbatim} 
\usepackage{enumitem} 
\nocopyright

\usepackage{todonotes}

\title{Balancing Efficiency and Coverage in Human-Robot Dialogue Collection}

\author[1]{Matthew Marge}
\author[1]{Claire Bonial}
\author[1]{Stephanie M. Lukin}
\author[1]{Cory J. Hayes}
\author[1]{Ashley Foots}
\author[2]{\authorcr Ron Artstein}
\author[1]{ Cassidy Henry}
\author[1]{Kimberly A. Pollard}
\author[2]{Carla Gordon}
\author[3]{Felix Gervits}
\author[2]{\authorcr Anton Leuski}
\author[1]{Susan G. Hill}
\author[1]{Clare R. Voss}
\author[2]{David Traum}

\affil[1]{U.S. Army Research Laboratory, Adelphi, MD 20783}
\affil[2]{USC Institute for Creative Technologies, Playa Vista, CA 90094}
\affil[3]{Tufts University, Medford, MA 02155}
\affil[ ]{\texttt {matthew.r.marge.civ@mail.mil }}

\begin{document}
\maketitle
\begin{abstract}
We describe a multi-phased Wizard-of-Oz approach
to collecting human-robot dialogue in a
collaborative search and navigation task.
The data is being used to train an initial automated robot dialogue system
to support collaborative exploration tasks.
In the first phase, a wizard freely typed robot
utterances to human participants.
For the second phase, this data was used to design a GUI
that includes buttons for the most common communications,
and templates for communications with varying parameters.
Comparison of the data gathered in these phases
show that the GUI enabled a faster pace of
dialogue while still maintaining high coverage of
suitable responses, enabling more efficient targeted
data collection, and improvements in natural language understanding
using GUI-collected data. As a promising first step
towards interactive learning, this work shows that our approach enables the collection of
useful training data
for navigation-based HRI tasks.
\end{abstract}

\section{Introduction}

Empirical data from  human-robot
interactions (HRI) can be used to enable robot dialogue systems to
interact naturally with people, and moreover support collaborative
tasks like interactive learning.
We present a multi-phased approach to automation of a robot dialogue
system, starting
with Wizard-of-Oz dialogue collection
and progressing towards full automation.
We explore this approach in the domain of collaborative
exploration between a human ``Commander'' and a
remotely located robot~\cite{Marge2016}.
We show that this multi-phased method, applied
successfully in virtual human research such as
SimCoach~\cite{simcoach2011}
and SimSensei~\cite{devault2014simsensei},
can be adapted to human-robot dialogue.

\begin{figure}[t!]
\centering
\includegraphics[width=3in]{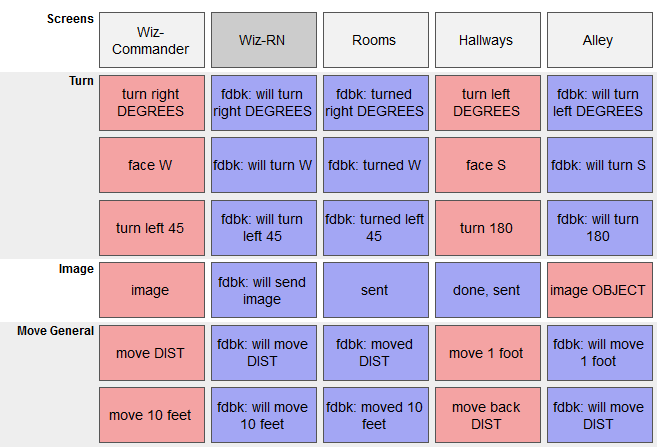}
\caption{Excerpt of a Wizard GUI for handling commands and composing replies to participants. Blue buttons reply to participant, while red ones route messages to an experimenter that teleoperates the robot.
CAPS indicate text-input slots.}
\label{dm-gui}
\vspace{-0.2in}
\end{figure}

An initial step is the collection of
human-robot dialogue to assess how humans would
naturally speak to
a robot in a collaborative exploration task,
including cases where
the robot would need to handle
confusing or insufficient instructions.
These data can serve as a source for establishing
the requirements for robot
dialogue capabilities, and also serve as training and evaluation
data for machine learning approaches to create these
capabilities.
Thus, the first phase in our approach (Experiment 1) is
an exploratory data collection of natural language dialogue,
where participants provide spoken instructions
to a robot,
and a wizard experimenter replies
as the robot via text responses in a chat window.
In Experiment 1, the wizard uses \textit{Free Response Mode}
to interact with participants by freely typing responses
following basic response guidelines.
The second phase (Experiment 2) uses
the collected corpus to design a click-button GUI
(see Figure~\ref{dm-gui}) featuring
the most common
communications, with some
parameters that can be entered
manually (e.g., ``turn right \textit{*135*} degrees'')~\cite{bonial2017laying}.
This form of interaction, \textit{Structured Response Mode},
has the potential benefit of eliciting dialogue
from participants more efficiently than typing, but it also
limits the wizard's communications to those in the GUI.
The data generated from the second phase can
provide a consistent distribution of dialogue
strategies for training a dialogue system and inform how
robots should respond with status updates and clarifications so that
tasks can be successful.

In this paper, we focus on exploring how the Free and Structured Response
Modes differ with respect to our goals of (1) eliciting
the natural diversity
of communication strategies in navigation tasks,
(2) simplifying
the data collection demands in order to
collect as much dialogue as possible within
the time constraints of experimental sessions, and (3)
creating a training dataset for natural language understanding (NLU)
and dialogue response generation algorithms.
We compare results with ten participants per experiment, resulting in
over thirteen hours of human-robot dialogue.
As a means to quantify differences in the data between
Free and Structured Response Modes, we pose the following
open questions.
(Q1):~How does the amount of data collected under Structured Response Mode
compare to Free Response Mode?
(Q2):~Is the human-robot communication productive
for the collaborative tasks?
(Q3):~Does Structured Response Mode achieve good coverage of the dialogue in the task domain by abstracting free text into buttons?
(Q4):~Does dialogue data collected with Structured Response Mode
result in better performance when training an automated
NLU and dialogue response generation component?

Our contributions are the following:
\begin{itemize}[noitemsep,nolistsep]
  \item Evaluation of a multi-phased Wizard-of-Oz approach to dialogue collection for HRI
  \item Annotations and measures for tracking dialogue efficiency and coverage
  \item Comparison of wizard response methods (Free and Structured), showing that Structured Response enables faster and more efficient targeted data collection while maintaining high coverage of suitable responses
  \item Improved understanding of robot's role in dialogue
  (i.e., issuing feedback, clarifications)
\end{itemize}

\section{Related Work}
\label{related-sec}

\subsection{Dialogue for HRI}
While natural language interaction
has been explored extensively in HRI \cite{mavridis2015review},
the primary focus, as described below, has been on analyzing
and determining strategies for one direction of communication
(human-to-robot or vice versa), but not both directions at the same time.

For human-to-robot communications, many approaches
follow the methodology of
\textit{corpus-based robotics} \cite{Bugmann04}, where
natural language in the form of route instructions are collected
from people (e.g., datasets
such as \textsc{Marco} \cite{Macmahon06walkthe} and the
TeamTalk corpus \cite{margeteamtalkcorpus}).
Computational approaches center around natural language
understanding (e.g.,~\cite{kruijff2010situated,williams2015going})
and symbol grounding methods that map language to symobolic
representations used for motion planning (e.g., Tellex et al.  \shortcite{tellex2011understanding};
Hemachandra et al.  \shortcite{hemachandra2015learning}).

Very limited effort has gone into developing
robot-to-human communications beyond researchers writing
the capabilities themselves. Some have made focused
efforts to understand how robots can
explain tasks \cite{foster2009evaluating}
or paths \cite{bohus2014directions,perera2016dynamic} to people in natural
language. Others have
developed computational methods to ask for clarification
about symbols \cite{deits2013clarifying} and to ask for help with tasks \cite{knepper2015recovering}.

Differences between a human and robot's internal
representation of an environment represent an instance
of the grounding problem \cite{clark_1996}
and must be resolved for grounding to occur.
Some have studied the nature of breakdowns
in human-robot communication
(e.g., Marge and Rudnicky \shortcite{Marge2015}),
while others have implemented real-time grounding
frameworks \cite{chai2016collaborative}. Several
dialogue interfaces have been developed for robots
(e.g., DIARC \cite{scheutz2018cognitive} and TeamTalk \cite{usarsimteamtalk}), but most rely on
handcrafted grammars or synthetic training data.
Our work builds upon previous research by investigating
empirical methods to human-robot \textit{dialogue} collection
(not unidirectional) that
strike a balance between eliciting
the naturally-occurring diversity of communication
strategies from participants and ensuring the data can be
patterned and tractable enough for training a dialogue system.

\subsection{Wizard-of-Oz (WoZ) Methodology}
WoZ design is a useful tool that has been widely adopted
by dialogue and HRI researchers because it allows
for low development costs and extremely malleable robot functionality.
WoZ has been used for handling natural
language since the early days of HRI \cite{riek2012wizard}
and dialogue \cite{fraser1991simulating}, and has been extended
to incorporate multi-wizard setups for
multimodal interfaces \cite{salber1993applying}
due to task complexities such as
supporting HRI
(e.g., Green, Huttenrauch, and Eklundh \shortcite{green2004applying}).
Wizards have also
played a role in collecting dialogue clarification
strategies \cite{passonneau2011embedded}.
Our work expands on these methods by addressing multimodal communication when the robot and human are not co-present, where information such as robot position, visual media, and dialogue would need to be exchanged.

The WoZ methodology has also been used successfully in fairly
open-domain tasks, for example a conversational assistant for
general-purpose information access which works by crowdsourcing
multiple wizards in real time \cite{Lasecki:2013:CCC:2501988.2502057},
or an agent for social conversation which uses
crowd-sourced wizards to expand its dialogue abilities
\cite{Kennedy-EtAl:2017:IVA}. These open-domain applications benefit
from access to many wizards with general human knowledge and limited
training. In contrast, a robot navigating a specific physical
environment requires fairly little knowledge (mostly about objects in
the environment), but we found that standing in for such a robot
requires substantial training \cite{marge-EtAl:2016:IVA}. Our work
therefore concentrates on the robot's navigation and communication
actions, rather than general knowledge.

Some criticisms of the WoZ approach have highlighted concerns about the validity of using human-human interaction disguised as a human-robot or human-agent interaction \cite{weiss2009userexp}, and successfully migrating a WoZ setup onto an autonomous robot \cite{breazeal2005effects}.
These concerns partially motivate the multi-phased approach
we have adopted from virtual human research \cite{devault2014simsensei}, but with extensions for situated dialogue where the robot must be aware of,
navigate, and refer to its surroundings while handling misunderstandings.

\section{Background}
\label{background-sec}

\subsection{Collaborative Exploration Domain}
\begin{figure}[t]
\includegraphics[width=3in]{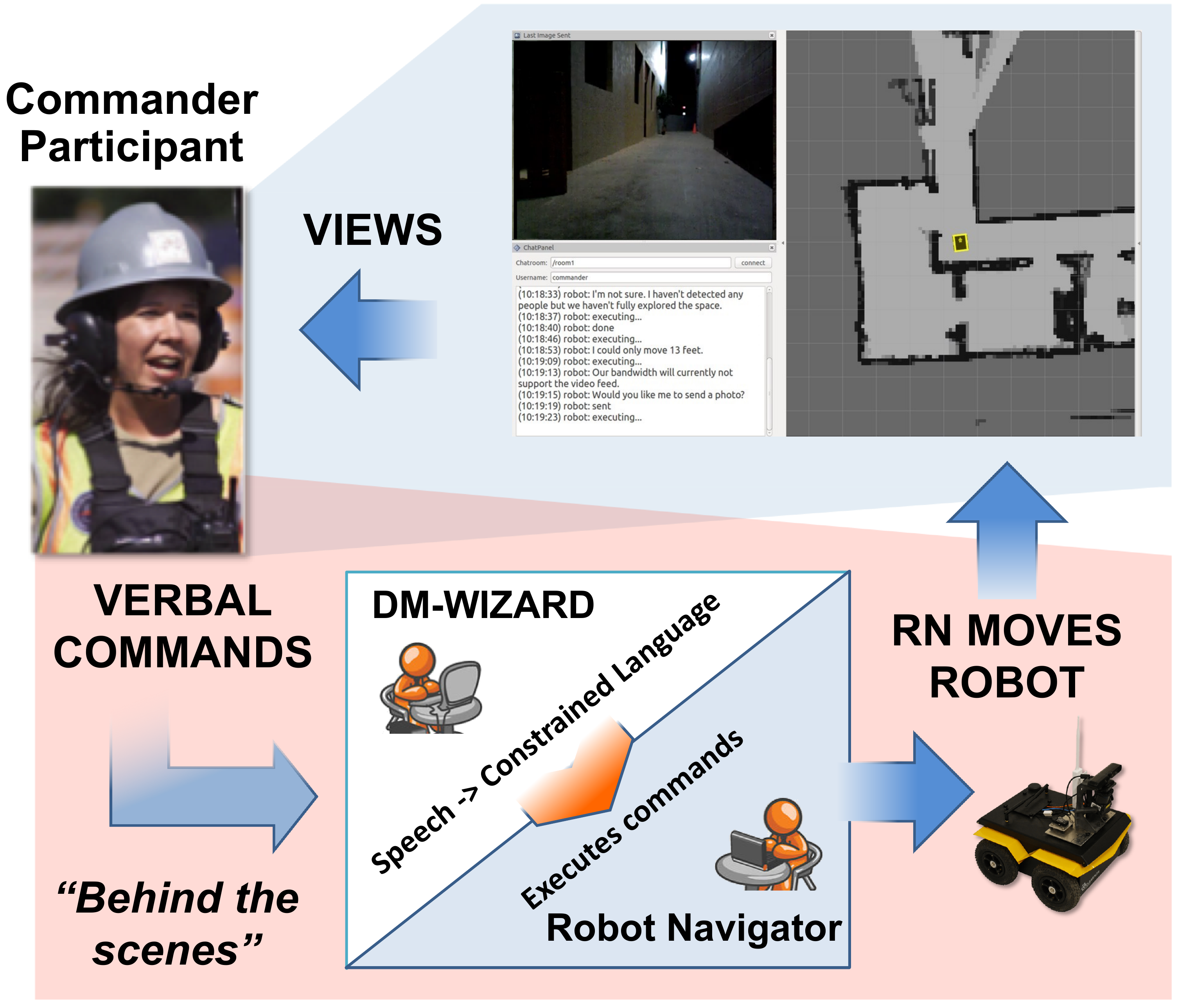}
\caption{\label{wozsetup} The Commander issues verbal
commands to the robot, whose capabilities are
performed by two wizards standing in for respective
abilities of dialogue and navigation.
}
\label{expt-layout}
\end{figure}

The domain testbed for our work is collaborative
exploration in a low-bandwidth environment.
This testbed mimics what can be found in reconnaissance and search-and-rescue operations---scenarios wherein a human
may verbally instruct a robot from a remote location.
The human ``Commander'' who instructs the robot
has specific goals for the exploration, such as locating
doors or types of objects in the physical
space, but is unable to directly act in or observe
this environment.
The Commander cannot directly teleoperate the robot,
but instead provides unconstrained spoken
instructions (e.g.,
``turn left 90 degrees,'' ``go through the doorway'')
to accomplish assigned tasks with the robot.
The Commander's knowledge of the environment is based
solely upon information streams provided by the robot
(see Figure~\ref{wozsetup}, upper right):
a LIDAR map of the area built up in real time
as the robot
moves, still images taken upon request, and text message
replies from the ``robot''.

\subsection{Multi-Wizard Setup}
\label{multi-wizard}
While the main focus of this paper is on using a wizard for
bootstrapping natural communication, in the initial phases of
this work we use a second wizard for robot navigation.
Each wizard takes the role of what we assume will ultimately be separate
modules in a fully autonomous robot.
In our setup, a Dialogue Manager Wizard (DM-Wizard) listens to
the Commander's speech and communicates directly
with him/her using a chat window to send text
status updates and requests for clarification. If
the instructions are executable in
the current context, then in another
chat window, the DM-Wizard passes a
simplified text instruction set to the Robot Navigator Wizard (RN), who
teleoperates the robot. When hearing robot status
updates directly from the RN, the DM-Wizard
communicates this information back to the Commander.
Figure~\ref{wozsetup} presents our setup.

\subsection{Multi-Phased Approach}
The multi-phased approach to
developing robot dialogue capabilities
consists of a series of Experiments:
In Experiment (Exp)-1, our goal was to elicit the full range
of communications that may arise in our domain.
To allow for this, the DM-Wizard spontaneously typed
responses ({\it Free Response Mode}, see Figure~\ref{tab:examples}) to
the Commander based on simple response/execution policy guidelines.
The guidelines identified the minimal requirements
for an executable instruction: instructions must contain both a clear
action and respective endpoint.
In Exp-1, the role of the DM-Wizard was
relatively burdensome, as the wizard not
only had to choose responses and compose messages to both the RN and to the
Commander, but also had to physically type the messages as quickly
and with as few typographical errors as possible.

 \begin{figure}[t]
 \centering
 \begin{small}
 \begin{tabular}{ |p{2.9in}| }
 \hline
 {\bf Free Response Mode (Experiment 1)} \\ \hline
 {\it P}: move forward \\
 {\it Robot (DM)}: You can tell me to move to something that
 	you see. \\ \hline

 {\it P}: continue to move down the hallway	 \\
 {\it Robot (DM)}: How far should I continue down the hallway?
 	Would you like me to reach something that you see
 	or go a certain distance? \\ \hline \hline

 {\bf Structured Response Mode (Experiment 2)} \\ \hline
 {\it P}: robot go forward	 \\
 {\it Robot (DM)}: I'm not sure where or when to stop
 	moving forward. You can tell me to move a certain
 	distance or to move to an object. \\ \hline

 {\it P}: can you move forward	\\
 {\it Robot (DM)}: I'm not sure where or when to stop
 	moving forward. You can tell me to move a certain
 	distance or to move to an object. \\ \hline

 \end{tabular}
 \end{small}
 \caption{Dialogues in Free Response Mode and Structured Response Mode between Commander Participants (P) and Robot (DM-Wizard).}
 \label{tab:examples}
 \vspace{-0.2in}
 \end{figure}

The Exp-1 data
were analyzed to extract a set of DM-Wizard messages and message templates
to strike a balance
between tractability for an autonomous robot and full coverage of
what humans were likely to say in the context of the task
domain, including how to elicit clarifications and recover from
problematic instructions. This set of
communications was incorporated in a
GUI (Figure~\ref{dm-gui}) for the DM-Wizard in Exp-2.
This {\it Structured Response Mode}
reduced the typing burden and much of the
composition burden for the DM-Wizard.
Figure~\ref{tab:examples} shows the greater uniformity in response policy for Structured Response Mode.

In the creation of the wizard interface, we considered the possibility
that there would be the need for responses to the Commander that had not
arisen in the data from Exp-1, thus not mapped to
a button in the interface. The GUI therefore includes buttons that
represent a general, non-understanding policy, which is used
in cases where no more specific response could have been
given. This might happen because the request was off-topic and there was
no proper response (e.g., ``are you male or female?''),
nonsensical in the current environment (e.g., ``turn 200 feet
left''), or outside capabilities in some way that had not
been encountered before.

\section{Data Collection Experiments}
\label{method-sec}

In both Experiments, the participant (Commander) performs a
collaborative search and navigation task with a robot
teammate to find objects in
a house-like environment as well as answer questions
about the environment.
The DM-Wizard role was kept constant by having the same experimenter
perform that role for both experiments.

\subsection{Experiment Design and Method}
Each participant first answered a questionnaire
to collect demographic information.
The participant was then seated at a
computer monitor, fitted with a headset microphone, and given
a push-to-talk button.
The participant was also given a list of the
robot's capabilities (see Appendix),
shown a photo of the robot, and was provided with a worksheet
listing the tasks and a pen for taking notes.
Participants viewed the interface shown
in Figure~\ref{expt-layout} (upper right), but
were unaware that the robot was
controlled by wizards.

\begin{figure*}[t]
\centering
\includegraphics[width=5.6in]{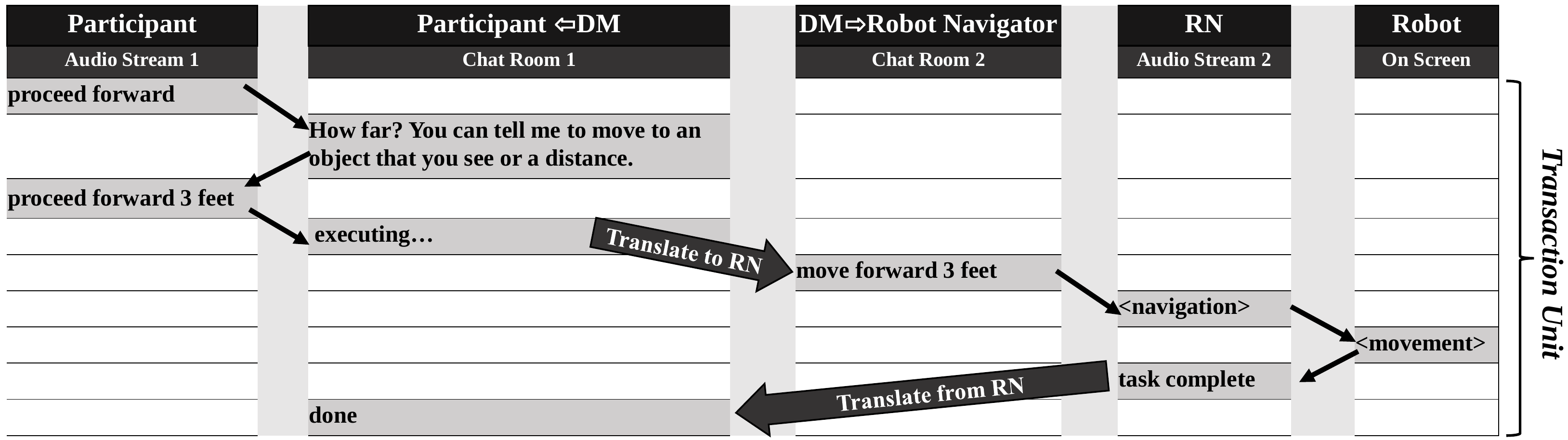}
\caption{Two wizards
manage the labor of robot intelligence. Dialogues
divide into \textit{transactions}
where a participant gives an instruction,
a \textit{Dialogue Manager} (DM-Wizard) decides how
to handle it, and the DM-Wizard
passes well-formed instructions
to a \textit{Robot Navigator} (RN) that moves the
robot.}
\label{open-ended}
\end{figure*}

Next, the participant completed
a training period in
which he/she was asked to perform navigation and search
tasks with the
robot in a remotely-located alley-like environment.
Once comfortable, the participant moved on to the two main
trials, in which the robot was placed in a new, house-like environment.
All environments were unfamiliar to participants.

Each trial had a different start
location within the house-like environment and a
different set of tasks, such as counting doorways, shovels,
or determining whether the space was recently occupied.
The order of the main trials was
counterbalanced across participants.
The main trials lasted until the participant reported
completion or 20 minutes, whichever occurred first.
We found that participants took the full
20 minutes.
No feedback was given as to
their performance of the tasks.

Ten people participated in each experiment.
People who participated in Exp-1 did not participate again in
Exp-2. In Exp-1, there were 8 male and 2 female participants,
and the mean age was 44
(min = 28, max = 58).
In Exp-2, there were 5 male and 5 female participants,
and the mean age was 42
(min = 18, max = 58).

\subsection{Corpus \& Annotations}
In addition to questionnaire data, we collected data from the experiments, including
speech of the participant and RN, text messages from DM-Wizard, and logs of all
robot images, maps, and navigation commands. The entire corpus
(training and main trials) consists
of recordings from 20 participants (approximately 20 hours
of audio; 3,573 utterances; 18,336 words). In addition to this raw
data, all speech was transcribed, and several kinds of annotation were
performed \cite{traum2018dialogue}, which we describe below.

\subsubsection{Dialogue Utterances}
We segment participant speech by separating it into individual
\textit{utterances}, which may range from single words to phrases (e.g.,~``Turn left 90 degrees and take a picture'' would segment as ``Turn left 90 degrees'' and ``and take a picture'').

\subsubsection{Dialogue Structure Annotation}
To follow information exchange
and assess the effectiveness of the
communication, we annotated the dialogue
using a dialogue structure annotation schema in which
sequential sets of utterances
involved with executing an
instruction are encoded as a \textit{Transaction Unit} (TU)~\cite{marge2017exploring}, and
each utterance is annotated for its structural role in the
exchange during that TU.
Figure~\ref{open-ended} shows the structure of a single TU;
there are four streams of communication: (1) the
participant speaking to the DM-Wizard, (2) the DM-Wizard
communicating with the
participant in text via a chat window, (3) the DM-Wizard
communicating with the the RN also in text via a chat window, and (4) the RN
speaking to the DM-Wizard.
Note that there
is no direct verbal communication path between
the participant and the RN---utterances
must be ``translated'' by the DM-Wizard.
The dialogue exchange in Figure~\ref{open-ended} depicts
a TU containing a \textit{Successful Instruction} (SI), that is,
a well-formed instruction
for which the RN reported successful execution.

\subsection{Measures}
We aim to assess differences in dialogue efficiency, dialogue coverage,
and training data utility between Free and Structured Response Modes.
We address four main questions (previously mentioned in the Introduction and summarized here):
First, we ask
whether more data is gathered per participant
when using the Structured Mode GUI (Q1; i.e.,
whether participants engage in more dialogue).
Second, we ask whether the human-robot communication is more
productive when
using the Structured Mode (Q2; i.e., more tasks completed).
Third, we ask whether the Structured Mode
GUI successfully achieves good coverage of the
dialogue used in the task (Q3).
Fourth, we ask if data collected will result in better
automated NLU performance (Q4; i.e., more useful training data).
For \textit{dialogue efficiency}, we measure both greater quantity of data (Q1), as well as higher task productivity (Q2) in terms of the
ability of the participant to effectively
communicate to the DM-Wizard,
who can then pass executable instructions to
the RN to navigate the space.
For \textit{dialogue coverage}, we tabulate occurrences
of wizard non-understanding under Structured Response Mode (Q3).
For measuring \textit{utility as training data}, we compute accuracy
at selecting gold standard dialogue responses with an NLU classifier (Q4).
The measures are described below.

\textbf{Dialogue Utterances and Words (Q1).}
A greater number of participant utterances and words indicates that a greater
sample of human and wizard language was collected,
and might suggest a productive data-gathering session.
We combine the number
of utterances from the participant and from the DM-Wizard to take
into account the
full sum of interactions between the two speakers.

\textbf{Dialogue Structure (Q2). }
Utterance count alone may or may not suggest a more productive human-robot
interaction in terms of
successful communication or task completion.
For example, if an initial instruction
did not contain sufficient information or was
misunderstood,
more utterances would be required
to clarify and repair the instruction,
leading to a more verbose, but not more productive, interaction.

To account for the potential correlation between more verbose instructions
and more unsuccessful interactions in terms of task completion, we
compute several metrics related to dialogue structure
annotations to assess dialogue efficiency.
A higher number of TUs corresponds with more instructions issued.
In addition, the number of TUs that
include a successful instruction (SI-TU) is a measure of communication effectiveness, namely
the participants' ability to work with
the ``robot'' (DM-Wizard) to issue a well-formed and executable instruction.

The metrics by themselves may be biased towards specific
instruction preferences or patterns.
A participant
may wait for the first instruction to be
completed before issuing another (e.g., ``Turn right 90 degrees''
and then after the first instruction is executed, ``Take a
picture,'' resulting in two TUs with one SI in each),
or may issue instructions in a group
(e.g. ``Turn right 90 degrees and take a picture,''
resulting in a single TU with two SIs).
To counter this potential bias in the SI-TU metric where
there may be multiple SIs within a TU, we
consider the total number of SIs independent of TUs.
Further, we compute an SI ratio per participant as the
number of TUs that contain an SI
divided by the total number of TUs (SI/TU ratio).
This metric ensures that no matter how many TUs were
issued, the percent of them that were well-formed and
executed will be normalized across participants
despite differences in instruction-giving preferences.

We note that the SI-TU and SI/TU metrics
count the entire TU as successful even if only part
had been completed before being abandoned.
The DM-Wizard will always engage in a clarification dialogue
with the Commander in the event that their instructions
that cannot be executed. However, the Commander may
abandon a TU in which the RN has only completed a subset
of the issued instructions. Since there is no way
for the DM-Wizard or RN to know \textit{a priori}
if the Commander will abandon the TU, we consider
these TUs successful---the RN accomplished the requested
tasks until the Commander decided to abandon their original request.

\textbf{GUI-Button Coverage (Q3).}
To measure coverage of the Structured Response Mode GUI,
we examine the use of the general, non-understanding
buttons and compute
the percent of utterances from the
DM-Wizard to
the participant that are of this type
(e.g., ``I'm not sure what you are asking me to do'').
These indicate that (1) there is no corresponding
button to pass the instruction to the RN
and/or (2) there is no way
to clarify the instruction in a manner that pinpoints
the specific problem.

\textbf{NLU Component Training Data (Q4).} Data from the
experiments can be used to train
machine learning algorithms for natural language understanding (NLU) and
response selection. The NLU component should map an incoming user utterance to
a representation that an automated dialogue manager can act upon;
while there are many possible structured representations that can fit
the task, we have not yet settled on a specific representation for our
future automated system. Therefore, as a proxy for a structured
representation, we use buttons from the DM-Wizard GUI. That is, we
test the ability of using data from the experiments to identify the
DM-Wizard's first
reaction to participant instructions in a held-out test
set (for example, relaying
the utterance to the RN or
asking the participant for clarification).

Using Exp-2 data for training and testing the NLU is straightforward,
because the DM-Wizard's reaction to each user utterance is a GUI
button press. In Exp-1, however, the DM-Wizard's reaction is free
text; in order to use Exp-1 data for training and testing the NLU, we
manually mapped each DM-Wizard text to the corresponding GUI button.
We held out one whole
dialogue from each experiment as test data, and used the remainder for
training.
Overall we had 33~test utterances and 595~training utterances
from Exp-1, and 52~test utterances and 977~training utterances from \mbox{Exp-2}.

We trained and tested different versions of the NLU component
using NPCEditor~\cite{LeuskiTraum2011},
a system that has been used to create classifiers for both structured
and free text natural language understanding.
We trained three versions of the NLU component, using Exp-1 data,
Exp-2 data, and the combined data; we then tested each one on the
Exp-1 test set, the Exp-2 test set, and the combined test set. Our
measure of performance is accuracy: a classifier response is
considered correct if it is identical to the DM-Wizard's action in the
test set. However, there are some cases of distinct but equivalent
DM-Wizard actions. For example, one of the test utterances is \emph{a
  hundred and eighty degrees to the right}, and one of the classifiers
mapped it to the action \emph{w-turn\_right\_180}; however, the
DM-Wizard's action in the test set was the equivalent action
\emph{w-turn\_180} (no direction specified). To reflect the classifier's correct
performance in cases such as this, we manually checked the output of
each classifier, and marked as correct cases where it chose an action
equivalent to the action in the test set.

\textbf{Questionnaire Measures.} Spatial ability has been found
to impact results on spoken language
use in spatial contexts~\cite{schober2009spatial}.
All participants
completed a Spatial Orientation Survey
to assess spatial
orientation ability~\cite{guilford1948guilford}.

\section{Results}
\label{results-sec}

All forty main trial dialogue sessions (20 minutes each;
two per participant)
were included in the between-subjects analysis.
We assessed parametric differences
of response mode
using a mixed-effects analysis of variance model (a
standard least squares regression with reduced
maximum likelihood~\cite{harville1977maximum}).
All measures were first assessed for normality
using the Shapiro-Wilk Goodness of Fit test.
For the analysis, the key independent
variable in the assessment
was response mode (Free or Structured).
Other
fixed effects included in the model were
age (given the skewed
nature of the participant pool towards older
participants) and scores on the spatial orientation
survey. Participant ID was included as a random effect
in the model.

\begin{table}
\small
\centering
\begin{tabular*}{\columnwidth}{@{\extracolsep{\fill}}lr@{}lr@{}l@{}}
\toprule
\raisebox{-1.99ex}[0pt][0pt]{Measure}
  & \multicolumn{2}{c}{Free} & \multicolumn{2}{c}{Structured} \\
\cmidrule{2-3}\cmidrule{4-5}
& Mean & Std err & Mean & Std err \\
\midrule
\# of Utterances**** & 128.6 & (5.12) & 190.9 & (7.51) \\
\# of Words & 378.3 & (21.2) & 317.15 & (14.84) \\
\# of TUs** & 34.4 & (2.47) & 46.3 & (2.93) \\
\# of SI-TUs** & 29.2 & (2.41) & 40.3 & (2.89) \\
\# of SIs** & 31 & (2.5) & 41 & (2.88) \\
SI/TU ratio & 0.83 & (0.019) & 0.87 & (0.018) \\
\bottomrule
\multicolumn{3}{l}{** $p < 0.01$; **** $p < 0.0001$}
\end{tabular*}
\caption{\label{tab:dialogue-efficiency} Dialogue Efficiency Measures per experiment, Avg.~Across Trials (N = 20 trials per experiment)}
\end{table}

\subsection{Dialogue Efficiency}
We analyzed dialogue efficiency by
measures associated with dialogue utterances,
participant words, TUs, and SIs
(Table~\ref{tab:dialogue-efficiency}).
Addressing (Q1), we tabulated the number of dialogue utterances
between the participant and DM-Wizard across
response mode.
Only response mode had a significant main effect
on total dialogue utterances (F[1, 16] = 28.9, p $<$ 0.0001).
We observed no significant main effects for response mode on
participant number of words.

Addressing (Q2), participants issued significantly more TUs
when the DM-Wizard used Structured Response Mode
compared to Free Response Mode. Only response mode
had a significant main effect on total TUs
(F[1, 16] = 11.8, p = 0.003).
Participants also completed significantly more
SI-TUs when the DM-Wizard used Structured Response Mode
compared to Free Response Mode. Only
response mode had a significant main effect on
total SI-TUs
(F[1, 16] = 10.9, p = 0.005).
The DM-Wizard also sent
more task completion
messages to the participant
(i.e., SIs)
with Structured over
Free Response Mode. Again, only response mode had a significant
main effect on total SIs
(F[1, 16] = 9.3, p = 0.008).
We observed no significant main effects for response mode
on SI-TU ratio.

\subsection{Dialogue Coverage}
Addressing (Q3), we measured coverage of
the DM-Wizard's ability in Structured Response Mode to
respond to participant instructions
by tabulating the number of TUs that did not contain a
non-understanding on the part of the DM-Wizard.
We observed 99\% coverage of responses via
the GUI in Structured Response Mode.
Only 1\% of TUs completed
during Structured Response Mode trials contained
a non-understanding (11 out of the total 926).
We found that 8 trials (out of 20 total) contained
a non-understanding; these trials had a TU
rate of non-understanding that ranged from 2-6\%.

\subsection{Utility as Training Data}
Addressing (Q4), performance of the three classifiers on the three
test sets is reported in Table~\ref{tab:classifier-accuracy}.
\begin{table}
\small
\begin{tabular*}{\columnwidth}{@{\extracolsep{\fill}}rcccccc}
\toprule
  & \multicolumn{6}{c}{Test data} \\
  \cmidrule{2-7}
  & \multicolumn{2}{c}{Exp-1} & \multicolumn{2}{c}{Exp-2}
  & \multicolumn{2}{c}{Total} \\
  & \multicolumn{2}{c}{\textit{(N=33)}} & \multicolumn{2}{c}{\textit{(N=52)}}
  & \multicolumn{2}{c}{\textit{(N=85)}} \\
  \cmidrule{2-3}\cmidrule{4-5}\cmidrule{6-7}
 Training data & Count  & \% & Count  & \% & Count  & \% \\
\midrule
Exp-1    & 30 & 91 & 46 & 88 & 76 & 89 \\
Exp-2    & 32 & 97 & 49 & 94 & 81 & 95 \\
Total  & 30 & 91 & 47 & 90 & 77 & 91 \\
\bottomrule
\end{tabular*}
\caption{\label{tab:classifier-accuracy}NLU classifier accuracy
for different training data sizes on the dialogue response generation
task. There were 595 training utterances from Exp-1 and 977
training utterances from Exp-2.}
\end{table}
We note that the accuracies are all fairly high, ranging from 88\%
to~97\%, demonstrating that data from the experiments is useful for
training an automated NLU component. The test data from Exp-1
consistently results in higher accuracies, suggesting that it's
probably an easier test set. As for the training sets, we note that
the best results for all test sets come from training on Exp-2 data
alone---better even than training on the combined data.

\section{Discussion}
\label{discussion-sec}

The results
show that this multi-phased approach holds promise
for collecting efficient human-robot dialogue data
to be used to achieve the goal of
autonomous conversational robots.
Thus far, we have found that Structured
Response Mode, which
required the development
of a GUI based on previously collected Free Response Mode data,
supports efficient data collection.
Structured Response Mode
enabled participants to
engage in more dialogue and issue more
executable instructions in the same
duration of the experiment, answering (Q1) and (Q2),
as measured by the reported utterance, TU, and SI measures.

While there are more TUs, SIs, and SI-TUs in Exp-2,
the proportion of them is the same as in Exp-1; this suggests that we do not suffer
a quality-loss between experiments. With more instructions issued, there are
potentially more opportunities for the participant to issue instructions
that could fail; yet we observe a sustained
quality of instruction-giving in Exp-2.

\begin{figure}[!t]
\centering
\begin{small}
\begin{tabular}{ |p{2.9in}| }
\hline
{\bf Confusing Instructions} \\ \hline
{\it P}: turn about a foot to your left	 \\
{\it P}: take a photo 	 \\
{\it P}: and a foot to your right 	 \\
{\it P}: take a photo 	 \\
{\it DM}: Hmm, I'm not sure what you would like me to do. \\
{\it DM}: Can you describe it another way?  \\ \hline
{\bf Novel Requests} \\ \hline
{\it P}: how far away are you from the object in front of you \\
{\it DM}: I'm not sure \\ \hline

\end{tabular}
\end{small}
\caption{Dialogues with low coverage in Structured Response Mode\label{tab:failedcov}}
\vspace{-0.2in}
\end{figure}

We also found that by using only the
human-robot dialogue collected in Exp-1 (10 participants), we could
build a GUI that supported reliable coverage of
natural language dialogue in the collaborative
exploration domain, answering (Q3). As measured by the number of
general non-understanding strategies initiated by the DM-Wizard,
very few situations could not be handled by the GUI.
While this may be in part due to the
restricted domain of navigation instructions, we note
that participants received no prior examples
from experimenters on how to formulate
instructions to the robot; they
used what they felt were good
instructions
based on their own intuition.

Although there was good coverage, we note
two instruction types in the 1\% of instructions that did not translate to
reliable responses in the GUI: instructions where the
DM-Wizard was genuinely uncertain of what action is being requested
(e.g., how should ``turn a foot'' be interpreted as degrees
of rotation)
and a novel type of request for something outside of the robot capabilities.
Examples can be found in Figure~\ref{tab:failedcov}.
The novel requests may have occurred in Exp-1, but not often enough
to have dedicated buttons in Exp-2 addressing them.
Further investigation is merited in this area.

Regarding (Q4), we have shown that the data collected in the
experiments is useful for training an automated NLU component, and
that Exp-2 resulted in higher quality data for training the
classifier.
This could be due to the
modality of using a GUI as opposed to free text, or possibly to the
fact that the DM-Wizard in Exp-2 was more experienced than in Exp-1.

The data collected in Structured Response Mode will
be particularly helpful for developing a future
robot dialogue system for several reasons.
First, more utterances were collected per trial in Structured
Response Mode than in the Free
Response Mode; this efficiency is important given the high cost of
collecting training data.
Second, the structured responses by the DM-Wizard provide a natural
classification of the corresponding participant utterances; this
``annotation through interaction'' will be helpful for training the
language understanding components, as shown by our initial tests on
classifier performance.
However, we note that Free Response Mode is essential
to the data collection process: the Structured Response
Mode using the GUI would not have been possible without
the bootstrapped dialogue data from Free Response.
Long-term, the solution for tractable data collection is
to move towards structured data elicitation.

\subsection{Qualitative Lessons Learned}
Based on both experiments, we found that speed and responsiveness
at processing dialogue data
are important for approaching a more realistic
and natural pace of dialogue. Structured
Response Mode
allows the participant
to complete more instructions when interacting with the DM-Wizard.
We also found that the fast-paced nature of the dialogue
requires simple messages to be sent
to the participant (e.g., ``processing\dots'')
to hold the conversational floor
while the DM-Wizard decides what to do next.
Feedback of this nature has the benefit of preventing situations
where the participant issues a command, receives
no response over a certain period of time, assumes
something went wrong, and issues another command.

We found that near-complete coverage of the
language in this domain was made possible by including
the following types of button categories in the GUI:
(1)~fixed buttons for common instructions and
clarifications,
(2)~slightly generalized buttons
(e.g., referring to ``which one?'' instead
of ``which cone?'') for less common referents,
(3)~flexible templates with slots for less
common metric references and
descriptions (e.g.,~``I see\dots''), and
(4)~very general non-understanding responses
for things that cannot be handled with other
buttons sensibly
(e.g.,~``I'm not sure what you're asking me to do\dots'').
A mix of templating and fixed buttons helped with
GUI efficiency as well: templatic when needed,
but these take longer, while fixed buttons
can generate quick replies. However, fixed buttons alone cannot
provide full coverage.

\subsection{Design and Research Implications}
The results we presented provide strong support
for a
systematic, data-driven approach that
feeds free response data from one series
of human-robot dialogue collection runs forward into
a structured GUI that allowed participants
to provide more executable instructions
than with the traditional free response
approach. At the same time, high coverage of
a navigation domain can be achieved with a fairly
limited number of participant sessions.
While the WoZ method
is often used to simulate NLU in order to understand a
phenomena, in this work it is used to provide a bootstrapped
dataset that can be used to train
a dialogue system.

The interface design and eventual autonomous behaviors
are driven directly by Wizard-of-Oz data collection,
as opposed to researchers predicting what users want, or creating
synthetic training data, as is common in traditional
dialogue systems research. This approach works to address
the ``cold start'' problem---what data
do you use to start training a system?---with a dataset
that approximates interaction with an idealized automated system
(i.e., wizards).

\section{Conclusions and Future Work}
\label{conclusion-sec}

We present a methodology for building a framework
of natural communication between humans and
robots. We described a novel multi-phased plan to achieve
this goal for HRI, the first two phases of which are complete:
an initial phase with
a wizard that manually typed
natural language responses, and a second
phase in which the wizard used a GUI designed
from the data collected in the first phase.
We developed an annotation scheme that became the
basis for tracking dialogue efficiency.
Results show that the GUI
enabled a faster pace of dialogue with more task completions;
all while maintaining high
coverage of suitable responses.
Robot status updates and clarifications could be generated quickly.
Moreover, data
collected with the GUI led to improved
performance on an automated natural language understanding
 classifier trained on the data.

The next step in our process will be to introduce simulation of both
the physical environment and robot, in order to collect data more
rapidly and safely validate the automated robot functions before
returning to the physical environment with a fully automated robot.
With an initial system trained from the
early experiments, we will
explore interactive learning approaches (e.g., one-shot learning)
about novel objects in the robot's surroundings that aren't observed
in the training data. We will leverage observed policies in the training
data for clarifying descriptions of objects.
Another opportunity for future work is to explore a semi-structured
approach that provides both the GUI and a free response text box
to generate responses.
Finally, the data and annotations collected as part of this study
represent a set of situations, natural language,
and robot sensory data that can be used to benefit
the broader research community.
Much of this data is planned to be publicly available
in the next year.

\section*{Acknowledgments}

This research was sponsored by the U.S. Army Research Laboratory.
The authors would like to thank Brendan Byrne, Taylor Cassidy,
A. William Evans, Anya Hee, Reginald Hobbs, Su Lei, and Douglas Summers-Stay
for their past contributions to this project,
and the anonymous reviewers for their helpful comments.

\section*{Appendix}
\appendix

\section{Robot Capabilities}
\label{sec:capabilities}
These are, verbatim, the capabilities provided on a sheet to study participants: \\
\noindent ``The robot can take a photo of what it sees when you ask.
The robot has certain capabilities, but cannot perform these tasks on its own. The robot and you
will act as a team.

Robot capabilities are:
\begin{itemize}[noitemsep,nolistsep]
\item Robot listens to verbal instructions from you.
\item Robot responds in this text box \emph{(Experimenter points to instant messenger box on screen)} or by taking action
\item Robot will avoid obstacles
\item Robot can take photos directly in front of it when you give it a verbal instruction
\item Robot will know what some objects are, but not all objects
\item Robot also knows:
\begin{itemize}
\item Intrinsic properties like color and size of objects in the environment
\item Proximity of objects like where objects are relative to itself and to other objects
\item A range of spatial terms like to the right of, in front of, cardinal directions like N, S
\item History: the Robot remembers places it has been
\end{itemize}
\item Robot doesn't have arms and it cannot manipulate objects or interact with
its environment except for moving throughout the environment
\item Robot cannot go through closed doors and it cannot open doors,
but it can go through doorways that are already open
\item Robot can only see about knee height ($\sim$ 1.5 feet)."
\end{itemize}

\bibliography{ai-hri2018-efficiency}
\bibliographystyle{aaai}
\end{document}